\documentclass[runningheads]{llncs}
\usepackage[T1]{fontenc}
\usepackage{graphicx}
\usepackage{amsmath}
\usepackage{theorem}
\usepackage{booktabs}
\usepackage{multirow}

\begin{document}
\title{CatalogAgent: A Supervisor-mediated Self-Learning System Enabling Context Engineering for GenAI Models}
\titlerunning{CatalogAgent: Self-Learning Enabling Context Engineering}

\author{Zhu Cheng\textsuperscript{*} \and Zhenming Wang \and Yu Tang \and Dan Liu \and Bryan Zhang \and Athanasios N. Nikolakopoulos \and Pranav Souri Itabada \and Jing Zhang \and Chih-Chi Chou \and Peng Gao \and  Fatemeh Mansoori \and Bharat Bojja \and Sarath Chander \and Sameer Thombare \and \\ Umit Batur \and Tarik Arici\textsuperscript{*}}
\authorrunning{Zhu Cheng et al.}
\institute{Amazon, Seattle, USA\\
\textsuperscript{*}Corresponding authors: \email{zzcheng@amazon.com}, \email{aricit@amazon.com}}

\maketitle              
\begin{abstract}
Product catalogs are the backbone of e-commerce sites, yet a large number of structured attributes (SAs) — such as material, color, and shape often have missing values. Typically, SA values are extracted from product information, including titles and descriptions. While LLM-based generator-evaluator frameworks have demonstrated effectiveness for SA prediction — where an LLM generates SA values and another evaluates them — they face challenges when the Generator and Evaluator produce conflicting outputs, as either component can make mistakes. We introduce \texttt{CatalogAgent}, a novel agentic system that continuously improves Generator and Evaluator models for e-commerce catalog enrichment. When disagreements arise from (1) internal conflicts between the LLM-based Generator and Evaluator, or (2) external feedback from sellers on LLM outputs, a Supervisor Agent intervenes to mediate these conflicts and make final decisions. The system also incorporates a Memory Base and a Memory Summarizer that stores Supervisor Agent activities from individual cases and aggregates patterns into learnings. These learnings are fed back to the worker Generator and Evaluator LLMs, enabling self-improvement without human intervention. Through context engineering — injecting learnings and insights into worker LLMs' contexts — the system successfully transfers the Supervisor's capabilities to the Generator and Evaluator, improving their performance by 15.24\% and 13.98\%, respectively. Our experiments demonstrate a new paradigm of Supervisor Agent-mediated self-learning systems for improving generative AI model accuracy.

\end{abstract}
%
%
%


\section{Introduction}
Catalog enrichment plays a vital role in simplifying seller listing processes and enhancing customer shopping experiences in e-commerce. High-quality catalogs serve as crucial resources for customers and sellers.


A product consists of unstructured elements such as titles, descriptions, and images, and structured attributes (SAs) such as \textit{material}, \textit{color}, and \textit{shape}. Each product belongs to a product type (PT), which defines the set of SAs applicable to that product. For example, the PT Shirt defines SAs such as \textit{shirt form type}, \textit{material}, \textit{size}. We refer to a PT and one of its SAs as a PT-Structured Attribute pair (PT-A), and a specific product combined with one of its SAs as a Product-Attribute pair — the basic unit of a single enrichment task. Catalog enrichment is the process of filling in missing SA values in product catalogs. Major challenges include accurately predicting missing SA values from unstructured elements and verifying existing SA values against those elements to detect errors.

Large Language Models (LLMs) have gained significant traction for predicting or evaluating SA values \cite{nikolakopoulos2023sage,huang2025attributeforge,satyadharma2025autoprompt,gao2026cascadeagent}. However, several challenges remain: (1) It remains non-trivial to determine accurate SA values from unstructured elements, particularly given the diversity of SAs across different PTs. (2) With a vast set of SAs spanning a large volume of PTs in e-commerce catalogs, it is challenging to detect high-level error patterns or issue clusters (such as PT-level errors) that could simplify the enrichment problem and guide effective solutions. (3) Production-scale catalog enrichment requires lightweight, cost-effective LLMs that can process millions of products efficiently, making systematic context engineering essential for maximizing model performance without sacrificing scalability.

\begin{figure}[t]
\center
        \includegraphics[width=\columnwidth]{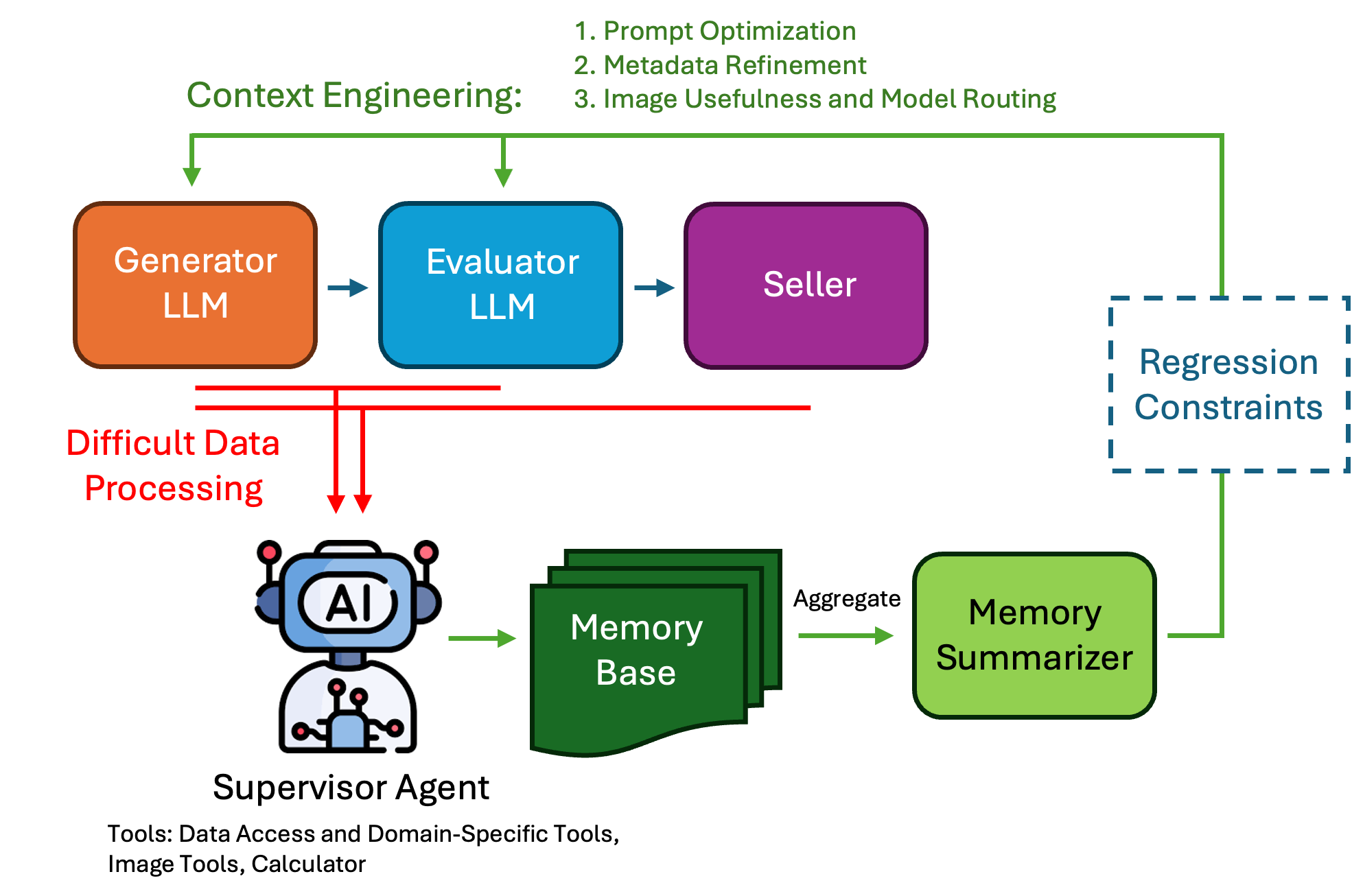}
    \caption{Overview of CatalogAgent system. The Supervisor Agent intervenes when disagreements arise — either between the Generator and Evaluator, or between models and Sellers. Memory Base stores individual case learnings from Supervisor investigations, and the Memory Summarizer aggregates them into generalizable insights that are fed back to the Generator and Evaluator through context engineering, forming a closed-loop self-learning cycle. A Regression Constraints module validates that updates do not degrade performance before deployment.
    }
    \label{fig:overall_figure}
\end{figure}

To address these challenges, we propose \texttt{CatalogAgent}, a novel Supervisor Agent-mediated self-learning system that performs \textbf{systematic context engineering} for worker Generator and Evaluator LLMs (Figure~\ref{fig:overall_figure}). A \textbf{Generator} infers specific SA values from product title, descriptions, and structured data — for example, determining \textit{battery type}, \textit{material composition}, or \textit{compatibility features}. An \textbf{Evaluator} verifies whether these values are accurate based on available evidence, creating an internal verification system. Both the Generator and Evaluator are designed as simple, efficient LLMs suitable for large-scale deployment: the Generator serves as the primary prediction model for attribute values, while the Evaluator acts as a quality control model to detect attribute value errors in catalog. This \textit{LLM-as-a-judge} paradigm has gained traction as a practical and cost-effective alternative to human evaluation, enabling judgments that surpass traditional matching or embedding-based methods \cite{li2025llmjudge,gu2024llmjudge}.

When both models agree on an attribute value, we have higher confidence in its accuracy. When they disagree, a \textbf{Supervisor} agent equipped with advanced reasoning capabilities and external tools arbitrates conflicts and determines the correct value. Such disagreements signal potential error patterns for specific structured attribute groups or product types, creating valuable learning opportunities for model improvement. A \textbf{Memory Base} stores these agent activities, and a \textbf{Memory Summarizer} extracts and aggregates learnings from this repository. These insights are validated through regression testing to ensure no degradation on existing performance, then fed back to the Generator and Evaluator models for targeted improvements through three types of context engineering: (1) prompt optimization to improve instruction clarity, (2) metadata refinement to identify incomplete value enumerations and propose additional valid values, and (3) image usefulness determination to identify which product images are most informative for specific PT-As.

This framework enables our system to autonomously monitor operations and enhance model quality through self-learning without human intervention. By engineering the right context — prompts, metadata, and input modalities — we can fully unleash the potential of lightweight worker LLMs. Moreover, the system is highly extensible and can incorporate additional signals from sellers or customers. For example, if sellers provide SA values that differ from the model-approved value, it may indicate a fundamental misunderstanding in a PT or its attribute definitions. Our system leverages such signals to autonomously self-improve.

\noindent \textbf{Contributions}: In this study, we introduce \texttt{CatalogAgent} — a novel supervisor agent-mediated self-learning system (Figure~\ref{fig:overall_figure}). The system autonomously monitors individual model operations and enhances attribute value generation quality through systematic context engineering without human intervention. Our contributions are three-fold: \vspace{-0.5em}                               

\begin{itemize} 
\item We introduce a new paradigm — an agent-mediated self-learning system that achieves high accuracy on attribute prediction and evaluation tasks while providing context engineering feedback to Generator, Evaluator, and Seller in a self-learning manner.

\item We present novel Memory Base and Memory Summarizer modules that create a self-improving feedback loop, capturing agent interactions and continuously feeding these learnings back to enhance the worker Generator and Evaluator models through context engineering.

\item We propose a unique approach for model improvement by mining difficult cases in catalog data between Generator and Evaluator, Model and Seller. This approach efficiently identifies specific areas requiring context engineering, rather than applying broad untargeted optimizations.
\end{itemize}

\vspace{-1em}
\section{CatalogAgent: A Supervisor-mediated Self-Learning System}
\subsection{System Overview}

Our system builds upon existing infrastructure where lightweight Generator and Evaluator LLMs process catalog enrichment at scale. These worker modules are standard LLMs optimized for high-volume batch processing.

We identify two critical signals that reveal model limitations: internal disagreements between Generator and Evaluator, and external conflicts where Sellers challenge the system-approved values. These challenging scenarios present opportunities for systematic improvement. To address these, we propose \texttt{CatalogAgent}, a novel Supervisor Agent-mediated self-learning system that performs systematic context engineering for Generator and Evaluator LLMs. \texttt{CatalogAgent} transforms difficult cases into continuous improvements by capturing failure signals, diagnosing root causes via the Supervisor's investigation, extracting generalizable patterns and insights, and injecting enhanced contexts back into worker models (Figure~\ref{fig:overall_figure}). A theoretical analysis of the Supervisor's impact on catalog precision and accuracy is provided in Appendix~\ref{sec:appendix}.

\vspace{-1em}
\subsection{Supervisor Agent as Investigator and Arbitrator}

The Supervisor Module operates as a high-reasoning agent that investigates difficult disagreement cases to generate granular, actionable insights. It arbitrates two types of conflicts: (1) \textit{Generator-Evaluator disagreement}: When the Generator predicts an attribute value that the Evaluator rejects, the Supervisor analyzes both models' reasoning chains to determine the correct attribute value and diagnoses which model erred and why, producing individual case level learnings. (2) \textit{Model-Seller disagreement}: When seller feedback conflicts with model-approved values, the Supervisor investigates the case to determine the accurate attribute value. In both scenarios, the Supervisor not only resolves the conflict but also generates insights that explain the root cause of errors and propose prompt refinements for Generator and Evaluator (Figure~\ref{fig:overall_figure}).

The Supervisor Agent employs ReAct methodology~\cite{yao2022react} for iterative, tool-augmented reasoning, enabling multi-step investigations rather than one-shot predictions. The investigation process is supported by a comprehensive toolbox that provides multiple capabilities: Data Access Tools retrieve catalog metadata and attribute schemas; Domain-Specific Tools include category-specific validators (e.g., size charts for Apparel), image analysis that extracts and verifies visual attributes from product images. The modular design enables rapid integration of new feedback signals and tools, allowing the Supervisor to continuously expand its capabilities as new data sources and validation methods become available.

\subsection{Memory Base: the Structured Foundation of Self-Learning}

The system's self-learning capabilities are enabled by the Memory Base and Memory Summarizer modules. The Memory Base stores Supervisor Agent activities from processing individual cases using a carefully designed schema (Table~\ref{table-agent-activity-schema-design-main}).


\noindent \textbf{Structured Schema for Context Engineering.} The Supervisor Agent follows an output schema that captures three categories of insights for engineering better contexts for the worker Generator and Evaluator models.

\textit{(1) Prompt Improvements} identify specific failure modes in the Generator and Evaluator. For instance, if the Evaluator consistently rejects accurate Generator predictions due to overly strict evidence requirements, the system updates the Evaluator's prompt to relax validation criteria for that PT-A.

\textit{(2) Metadata Refinement} addresses gaps in attribute definitions that cause systematic errors. For example, if \texttt{button down shirt} is a valid \textit{shirt form type} but absent from the predefined value set, neither the Generator nor the Evaluator can handle it correctly. Similarly, if an attribute definition is ambiguous — such as whether \textit{fit type} describes a garment's relationship to the body or its length — models will produce inconsistent predictions. Metadata refinement addresses these issues by expanding value sets, clarifying definitions, or updating rules.



\textit{(3) Image Usefulness Analysis} captures which images proved useful during the Supervisor's investigation for each product. For example, when investigating a handbag's inner capacity attribute, the schema records that secondary images showing the bag opened from above were necessary, whereas for a shirt's form type attribute, only the main image was needed. These individual case records accumulate in the Memory Base, providing the foundation for subsequent aggregation to identify systematic image usefulness patterns at the PT-A level.

\begin{table}[t]
\centering
\small
\caption{Supervisor Agent Output Schema for Context Engineering.}
\label{table-agent-activity-schema-design-main}
\begin{tabular}{lp{0.5\columnwidth}}
\toprule
\textbf{Field} & \textbf{Description} \\
\midrule
\multicolumn{2}{l}{\textit{Case Identification}} \\
\texttt{product\_id} & Product identifier \\
\texttt{product\_type} & Product Type \\
\texttt{attribute\_name} & Attribute being analyzed \\
\texttt{current\_value} & Existing attribute value \\
\texttt{disagreement\_type} & Generator-Evaluator or Model-Seller \\
\midrule
\multicolumn{2}{l}{\textit{Arbitration Result}} \\
\texttt{correct\_value} & Supervisor-determined attribute value \\
\texttt{generator\_correct} & Boolean: Whether Generator was correct \\
\texttt{evaluator\_correct} & Boolean: Whether Evaluator was correct \\
\midrule
\multicolumn{2}{l}{\textit{Structured Schema for Context Engineering}} \\
\multicolumn{2}{l}{\textbf{(1) Prompt Improvements}} \\
\texttt{instructions\_to\_generator} & List of prompt refinements if G erred \\
\texttt{instructions\_to\_evaluator} & List of prompt refinements if E erred \\
\multicolumn{2}{l}{\textbf{(2) Metadata Refinement}} \\
\texttt{metadata\_clarification} & Clarifications to attribute definition \\
\multicolumn{2}{l}{\textbf{(3) Image Usefulness Analysis}} \\
\texttt{image\_analysis.useful} & Boolean: Whether images are predictive for this attribute \newline List of image indices with predictive signal \\

\bottomrule
\end{tabular}
\end{table}

\subsection{Memory Summarizer: From Fine-grained Insights to Generalizable Learnings}
The Memory Summarizer processes records from the Memory Base to aggregate granular, case-specific insights into generalizable learnings at appropriate aggregation levels. For each PT-A pair, the Memory Summarizer retrieves all individual product-level suggestions from Supervisor investigations and identifies common patterns across failures at PT-A level. For example, for PT=Cellular Phone Case and Structured Attribute=\textit{model name}, the Memory Summarizer fetches individual suggestions such as ``be more cautious about assigning brand names as model names'', ``distinguish between pattern/design names and actual model names''. By analyzing these recurring failure modes across multiple products, the Memory Summarizer synthesizes a consolidated prompt instruction: ``When generating the model name attribute, focus on identifying a unique, specific identifier that distinguishes it within the manufacturer's lineup. Avoid using brand names, general product descriptions, color variants, pattern names, or compatibility information as model names. If no clear model name is provided, return Not Obtainable''. For prompt improvements and metadata refinements, the Memory Summarizer synthesizes common failure patterns into consolidated instructions and attribute definition updates.

The Memory Summarizer also aggregates individual image usefulness records to determine PT-A level patterns. When secondary images consistently proved necessary across cases — such as Handbag-\textit{inner capacity} — the configuration includes secondary images in the model inputs. Conversely, when only the main image proved useful — such as Shirt-\textit{form type} — the configuration optimizes to use only the main image, reducing input tokens and computational costs at scale.




\subsection{Context Engineering under Regression Constraints}
\label{sec:regression-constraints}

While context engineering through prompt improvements and metadata refinements targets specific difficult failure cases, it is critical to ensure that these targeted new contexts do not degrade the Generator and Evaluator models' performance on the general population of well-understood cases. We express this dual objective — reducing difficult disagreement cases while preserving general performance — as a constrained optimization problem. Let $\theta$ denote the configuration (including prompt text, metadata), and let $\theta_0$ represent the baseline configuration. We define:

\begin{itemize}
  \item $\mathcal{D}(\theta)$: Disagreement rate under configuration $\theta$, measured over Product-Attribute pairs where disagreements occur.
  \item $\mathcal{G}$: Golden dataset of Product-Attribute pairs with ground truth labels curated by human auditors, where $|\mathcal{G}| \ll |\mathcal{A}|$ and $\mathcal{A}$ represents all individual Product-Attribute prediction tasks in the catalog — each combination of a product and one of its PT-defined attributes constitutes a single task.
  \item $\mathcal{P}(\theta; \mathcal{G})$: Accuracy on the golden dataset $\mathcal{G}$ under configuration $\theta$.
\end{itemize}

The optimization objective seeks to minimize disagreement rates while ensuring non-degradation on the general population as represented by $\mathcal{G}$:

\begin{equation}
\begin{aligned}
\min_{\theta} \quad & \mathcal{D}(\theta) \\
\text{s.t.} \quad & \mathcal{P}(\theta; \mathcal{G}) \geq \mathcal{P}(\theta_0; \mathcal{G}) - \delta
& \theta \in \Theta
\end{aligned}
\end{equation}
where $\delta \geq 0$ is a small tolerance threshold for acceptable performance degradation, and $\Theta$ represents the feasible space of configurations.

For completeness, we note that the Generator and Evaluator employ distinct configurations $\theta_G$ and $\theta_E$ respectively. The joint optimization becomes:


\begin{equation}
\begin{aligned}
\min_{\theta_G, \theta_E} \quad & \mathcal{D}(\theta_G, \theta_E) \\
\text{s.t.} \quad & \mathcal{P}_G(\theta_G) \geq \mathcal{P}_G(\theta_G^0) - \delta_G \\
& \mathcal{P}_E(\theta_E) \geq \mathcal{P}_E(\theta_E^0) - \delta_E \\
& \theta_G \in \Theta_G, \; \theta_E \in \Theta_E
\end{aligned}
\end{equation}
where $\mathcal{D}(\theta_G, \theta_E)$ represents the disagreement rate under the joint configuration, $\mathcal{P}_G$ and $\mathcal{P}_E$ denote Generator and Evaluator accuracy on their respective golden datasets $\mathcal{G}_G$ and $\mathcal{G}_E$, with tolerance thresholds $\delta_G$ and $\delta_E$.

The constraint ensures that any new configuration must maintain near-equivalent or better performance on the golden dataset, preventing regression on well-understood cases. The golden dataset $\mathcal{G}$, while necessarily limited due to human annotation costs ($|\mathcal{G}| \ll |\mathcal{A}|$), serves as a representative sample of the general catalog population and acts as a safeguard against overfitting to disagreement cases. This formulation captures the fundamental trade-off in production ML systems: aggressively optimizing for newly discovered failure modes while preserving gains on previously solved problems.


\noindent \textbf{Regression Testing and Deployment.} The Regression Test Module validates that aggregated learnings from the Memory Summarizer do not degrade performance on the golden dataset $\mathcal{G}$. We focus exclusively on regression testing because the disagreement cases triggering learnings are, by definition, edge cases not covered in $\mathcal{G}$ — if they were present in $\mathcal{G}$, we would not observe these disagreements. Once validated, the improved prompt instructions and metadata refinements are integrated into the Generator and Evaluator configurations, completing the closed-loop learning cycle. The system then monitors online disagreement metrics $\mathcal{D}(\theta)$ in production as the primary indicator of improvement.

\vspace{-1em}
\section{Experiments and Results}
\label{sec:experiments}

\vspace{-0.6em}
\subsection{Dataset and Experimental Settings}
We identified 4.89 million Product-Attribute pairs where the Evaluator rejected the Generator's predictions. We randomly sampled 50k Product-Attribute pairs for the Supervisor to investigate and generate learnings, and a separate 50k for testing whether the resulting context engineering enhanced the worker LLMs. 
The PT-A distribution is long-tailed, with the leading PT-As being Shirt-\textit{shirt form type} and Cellular Phone Case-\textit{subject character}. 23k Product-Attribute pairs from seller feedback were used — these represent cases where sellers provided different values from those generated and approved by Generator and Evaluator. The Supervisor Agent is equipped with tools including a metadata lookup function, image analysis, calculator function, with access to product data, and attribute definitions. The worker Generator and Evaluator use Mistral-NeMo, a lightweight LLM suitable for high-volume processing. Supervisor Agent is powered by Claude Sonnet, chosen for its reasoning and tool-use capabilities required for multi-step investigations.


\vspace{-0.6em}
\subsection{Supervisor Agent as an Arbitrator} To resolve the conflict between Generator and Evaluator, Supervisor Agent monitors both models by analyzing their input and output. Acting as an arbitrator, the Supervisor Agent identifies true positives from each model and reveals domain-specific error patterns: Generator-dominant errors predominantly occur in the following PT-As: Shirt-\textit{shirt form type}, Cellular Phone Case-\textit{model name}. Evaluator-dominant errors are more common in \textit{material}, \textit{target gender} attributes (Table ~\ref{table-agent-separates-TP-FP}). This granular error attribution facilitates targeted improvements for Generator and Evaluator.

\begin{table}[h!]
  \caption{Error attribution in Generator-Evaluator disagreements mediated by Supervisor Agent.}
  \label{table-agent-separates-TP-FP}
  \centering
  \resizebox{1\columnwidth}{!}{
  \begin{tabular}{p{0.3\columnwidth}p{0.3\columnwidth}p{0.3\columnwidth}p{0.3\columnwidth}}
    \toprule
    Product Type & Attribute & Generator error (\%) & Evaluator error (\%) \\
    \midrule
    Shirt & shirt form type & 97.78 & 2.22     \\
    Cellular Phone Case & model name & 98.64 & 1.36     \\
    Shoe & model number & 94.87 & 5.13     \\
    Boot & material & 39.13 & 60.87     \\
    Shoe & material & 30.23 & 69.77     \\
    Shirt & target gender & 2.07 & 97.93     \\
    Shoe & shoe type & 48.78 & 51.22     \\
    Boot & boot form type & 62.22 & 37.78     \\
    Hat & fit type & 23.81 & 76.19     \\
    Shirt & top style & 56.25 & 43.75     \\
    \bottomrule
  \end{tabular}
  }
\end{table}

Agreement between Generator and Evaluator does not guarantee accurate attribute prediction. We discovered this through seller feedback that challenge the potentially incorrect values produced by models. The Supervisor Agent serves as a judge, analyzing both model-generated and seller-proposed values to determine a correct value. Our analysis revealed that models were typically the source of errors (Table ~\ref{table-agent-separates-TP-FP-ModelSeller}). For instance, in classifying \textit{closure type} for leggings, the presence of a small zippered pocket misled the LLMs to incorrectly label the \textit{closure type} as \texttt{zipper} instead of the correct value \texttt{pull-on} — an error that both Generator and Evaluator failed to catch. Interestingly, sellers were not always right either. Seller proposed values were incorrect in 76.60\%, 42.86\%, 48.98\% of the feedback on \textit{fit type}, \textit{water resistance level}, \textit{item form}, respectively (Table ~\ref{table-agent-separates-TP-FP-ModelSeller}). Sellers were confused between certain attributes. They incorrectly submitted value \texttt{25" Inseam} as \textit{fit type} of Pants, but rather attribute \textit{inseam length} is an appropriate destination. These insights help target specific areas for model improvement while identifying opportunities for seller education. Supervisor Agent's output schema (Table ~\ref{table-agent-activity-schema-design-main}) serves multiple purposes: it not only arbitrates errors (role-modeling for the Evaluator's assessment tasks), but also generates new attribute values after analyzing different conflicting view points, thereby acting as a role model for the Generator.

\begin{table}
  \caption{Error attribution in Model-Seller disagreements mediated by Supervisor.}
  \label{table-agent-separates-TP-FP-ModelSeller}
  \centering
  \resizebox{0.8\columnwidth}{!}{
  \begin{tabular}{p{0.4\columnwidth}p{0.3\columnwidth}p{0.3\columnwidth}}
    \toprule
    Attribute & Model error (\%) & Seller error (\%) \\
    \midrule
    care instructions & 81.08 & 16.22     \\
    fit type & 14.89 & 76.60     \\
    closure type & 91.67 & 4.17     \\
    material & 85.71 & 4.08     \\
    water resistance level & 53.06 & 42.86     \\
    directions & 91.67 & 8.33     \\
    model number & 87.76 & 10.20     \\
    recommended uses for product & 89.80 & 8.16     \\
    compatible devices & 71.43 & 20.41     \\
    number of items & 73.91 & 23.91   \\
    \bottomrule
  \end{tabular}
  }
\end{table}

\vspace{-3em}
\subsection{Supervisor Agent Performance Benchmark}

To validate the Supervisor Agent's effectiveness, we conducted human evaluations on both types of disagreement cases. For Generator-Evaluator disagreements, we randomly sampled 148 Product-Attribute pairs from cases where the Generator and Evaluator disagreed and compared the Supervisor's outputs against human assessment. The Supervisor achieved 97.26\% accuracy when performing the Generator's role (SA value generation) and 94.56\% accuracy when performing the Evaluator's role (SA value correctness assessment). For Model-Seller disagreements, human auditors evaluated 567 Product-Attribute pairs where sellers provided values differing from model-approved values. The Supervisor achieved 85.65\% precision and 94.28\% recall in in arbitrating between model-approved and seller-proposed values (Table~\ref{table-agent-accuracy-benchmark-against-human}). The Supervisor Agent's performance can be attributed to its comprehensive approach: it considers both Generator and Evaluator perspectives, employs detailed multi-step reasoning via ReAct, and breaks down complex investigations into manageable sub-tasks.

\vspace{-1em}
\begin{table}
    \caption{Supervisor Agent performance measured against human evaluation.}
    \label{table-agent-accuracy-benchmark-against-human}
    \centering
    \begin{tabular}{lc}
      \toprule
      \multicolumn{2}{c}{\textit{Generator-Evaluator disagreements (n=148)}} \\
      \midrule
      Generation Accuracy & 97.26\% \\
      Evaluation Accuracy & 94.56\% \\
      \midrule
      \multicolumn{2}{c}{\textit{Model-Seller disagreements (n=567)}} \\
      \midrule
      Precision & 85.65\% \\
      Recall & 94.28\% \\
      \bottomrule
    \end{tabular}
\end{table}


\vspace{-3em}
\subsection{Worker LLMs self-improve by incorporating aggregated learnings from Supervisor}
The Memory Summarizer and Regression Constraints modules are key to enabling self-improving capabilities of the system. It leverages stored agent activities from Memory Base to refine attribute metadata, improve Generator and Evaluator prompts, and determine LLM cascade configurations for various Product-Attribute tasks, routing each to the appropriate text-only or multimodal model with the optimal image inputs. This creates a self-learning loop where the worker Generator and Evaluator LLMs constantly evolve and improve based on accumulated insights and learnings from Supervisor Agent (Figure ~\ref{fig:overall_figure}).

\paragraph{\textbf{Learnings for metadata refinement.}}
One of the key improvement opportunities identified by the Supervisor Agent is metadata enhancement. Analysis of Generator-Evaluator disagreements revealed that many classification errors occur when the Generator must select from a constrained set of predefined values that lack appropriate options. For example, the attribute \textit{shirt form type} has predefined values such as \texttt{sport jersey}, \texttt{polo shirt}, and \texttt{t-shirt}, but lacks common styles like button down shirts, blouses, and henley shirts. When processing individual cases, the Supervisor Agent repeatedly observed that garments did not fit into any existing category, and systematically identified new values to add: \texttt{button down shirt}, \texttt{blouse}, \texttt{henley shirt}. These suggested additions undergo validation through a dedicated metadata agent before being integrated into the catalog schema.

%

As shown in Table ~\ref{table-metadata-update}, expanding the metadata values led to accuracy improvements of up to 5\% for certain PT-As. These results demonstrate how systematic analysis can surface critical metadata gaps that, when addressed through coordinated efforts of Supervisor and metadata agents, significantly enhance classification accuracy across both Generator and Evaluator.

\vspace{-1.4em}
\begin{table}
  \caption{Accuracy of generations from the worker LLM after metadata update.}
  \label{table-metadata-update}
  \centering
  \resizebox{0.95\columnwidth}{!}{
  \begin{tabular}{p{0.3\columnwidth}p{0.3\columnwidth}p{0.3\columnwidth}p{0.3\columnwidth}}
    \toprule
    Product Type & Attribute & Accuracy, Baseline (\%) & Accuracy, with metadata update (\%) \\
    \midrule
    Shirt & shirt form type & 87.95 & 92.54     \\
    Shoe & shoe type & 92.75 & 95.09     \\
    Boot & boot form type &74.47 & 79.80     \\
    Pants & pants form type & 87.69 & 90.35     \\
    Hat & hat form type & 93.57 & 94.44     \\
    \bottomrule
  \end{tabular}
  }
\end{table}

\vspace{-2em}
\paragraph{\textbf{Learnings for prompt improvement.}}
The schema of Supervisor Agent's output includes feedback messages from the Supervisor to Generator and Evaluator. Using Memory Summarizer, we compiled insights from individual cases handled by the supervisor. These consolidated learnings were then incorporated into the prompts for the Generator and Evaluator worker models. For example, in cases where a specific model name for a cellular phone case is not clearly identifiable, it should output Not Obtainable instead of attempting to create a model name from available information. Analyzing and aggregating from these agent activities led to a prompt update — ``When generating the \textit{model name} attribute, avoid using brand names, general product descriptions, color variants, pattern names, or compatibility information as model names, if no clear model name is provided, return Not Obtainable''. Memory Summarizer identified that the problem lies in the hallucinated nature of worker Generator LLM — it tends to fabricate phone case model names rather than admitting when one is not available. After incorporating prompt updates from Memory Summarizer, the worker Generator achieved up to 15.01\% improvement (Table \ref{table-learnings-prompt}).


\vspace{-1.2em}
\begin{table}
  \caption{Worker LLM improved from prompt injection by Supervisor Agent.}
  \label{table-learnings-prompt}
  \centering
  \resizebox{1\columnwidth}{!}{
  \begin{tabular}{p{0.3\columnwidth}p{0.3\columnwidth}p{0.3\columnwidth}p{0.3\columnwidth}}
    \toprule
    Product Type & Attribute & Accuracy, Baseline (\%) & Accuracy, with prompt update (\%) \\
    \midrule
    Cellular Phone Case & model name & 69.20 & 84.21     \\
    Shoe & model number & 65.39 & 78.80     \\
    Hat & material & 70.47 & 83.68     \\
    Rug & care instructions & 43.89 & 50.56     \\
    Cellular Phone Case & subject character & 85.85 & 86.87     \\
    \bottomrule
  \end{tabular}
  }
\end{table}


\vspace{-3em}
\paragraph{\textbf{Prompt improvement under regression constraints.}}
As formulated in Section~\ref{sec:regression-constraints}, each new configuration $\theta$ must satisfy the constraint $\mathcal{P}(\theta; \mathcal{G}) \geq \mathcal{P}(\theta_0; \mathcal{G}) - \delta$, ensuring that targeted context engineering does not degrade performance on the general population represented by $\mathcal{G}$. For each PT-A where the Memory Summarizer produces updated prompts with injected learnings, we compare precision and recall against the baseline configuration $\theta_0$ on the golden dataset. A prompt update is accepted if both metrics remain within the tolerance threshold $\delta$. As shown in Table~\ref{table-regression-test}, most PT-As pass regression testing, with several showing notable improvements — for example, Nail Polish-\textit{material type free} improved by +11.53\% in precision and +17.50\% in recall. A few PT-As exhibited degradation exceeding $\delta$; their prompt updates are rejected, preserving hard-won gains on previously solved cases. This validates the constrained optimization formulation: the system aggressively optimizes for newly discovered failure modes while safeguarding performance on well-understood cases.

\begin{table}[t]
    \caption{Regression testing results on the golden dataset. New prompts with context engineering feedback are validated against golden dataset on both precision and recall.}
    \label{table-regression-test}
    \centering
    \resizebox{\columnwidth}{!}{
    \begin{tabular}{llccccccc}
      \toprule
      Product Type & Attribute & Prec ($\theta_0$) & Prec ($\theta$) & $\Delta$ Prec & Rec ($\theta_0$) & Rec ($\theta$) & $\Delta$ Rec \\
      \midrule
      Cellular Phone Case & compatible devices & 78.57\% & 77.78\% & -0.79\% & 96.25\% & 96.25\% & 0.00\% \\
      Shorts & fit type & 28.07\% & 26.56\% & -1.51\% & 20.00\% & 21.25\% & +1.25\% \\
      Pants & care instructions & 46.32\% & 45.00\% & -1.32\% & 55.00\% & 56.25\% & +1.25\% \\
      Pants & fit type & 43.00\% & 44.00\% & +1.00\% & 53.75\% & 55.00\% & +1.25\% \\
      Essential Oil & product benefit & 94.94\% & 97.47\% & +2.53\% & 93.75\% & 96.25\% & +2.50\% \\
      Electronic Cable & connector type & 71.74\% & 71.28\% & -0.46\% & 82.50\% & 83.75\% & +1.25\% \\
      Nail Polish & material type free & 48.89\% & 60.42\% & +11.53\% & 55.00\% & 72.50\% & +17.50\% \\
      Portable Electronic Device Cover & compatible devices & 69.79\% & 71.13\% & +1.34\% & 83.75\% & 86.25\% & +2.50\% \\
      Pillow & fill material & 83.54\% & 77.78\% & -5.77\% & 83.54\% & 79.75\% & -3.80\% \\
      Necklace & metal type & 77.78\% & 83.78\% & +6.01\% & 70.00\% & 77.50\% & +7.50\% \\
      Headphones & noise control & 72.22\% & 68.60\% & -3.62\% & 81.25\% & 73.75\% & -7.50\% \\
      \bottomrule
    \end{tabular}
    }
\end{table}

\paragraph{\textbf{Learnings for image usefulness and LLM model selection.}}
By studying the patterns stored in Memory Base from individual agent activities, we can further improve how we utilize images more effectively. For example, for a Wall Art product, the main image only displays the product details and features, making it unclear whether the product is intended for indoor or outdoor use, whereas the 2nd - 4th images show the Wall Art product displayed in dining and living rooms, which suggests the \textit{indoor outdoor usage} to be indoor. Some PT-As are easily determined from the main image and text — like boot styles or shoe types. Specific details like a handbag interior material or its capacity to hold devices (attribute \textit{maximum compatible device size}) often require secondary non-main images that show close-ups, inside views, or size comparisons. For example, a secondary image might show a top-down view of an open handbag revealing its fabric lining (attribute \textit{inner material}). Certain PT-As benefit from the main image, whereas other PT-As benefit from non-main images (Table ~\ref{table-agent-learns-LLM-cascading-levels}). We further used the insights and activities from Supervisor Agent and Memory Summarizer to route each PT-A to the appropriate text-only or multimodal model with the optimal image inputs.



\begin{table}
    \caption{Top PT-As that benefit from predictive signals in main vs non-main image(s).}
    \label{table-agent-learns-LLM-cascading-levels}
    \centering
    \resizebox{1\columnwidth}{!}{
    \begin{tabular}{p{0.3\columnwidth}p{0.3\columnwidth}|p{0.3\columnwidth}p{0.4\columnwidth}}
        \toprule
        \multicolumn{2}{c|}{Predictive signals in \textbf{main} image} & \multicolumn{2}{c}{Predictive signals in \textbf{non-main} images} \\\midrule
        Boot & boot form type & Clock & operation mode     \\
        Hat & hat form type & Handbag & inner material     \\
        Pants & pants form type & Rug & water resistance level     \\
        Sweatshirt & target gender & Handbag & compatible device size maximum     \\
        Pants & fit type & Wall Art & frame material     \\
        Shirt & shirt form type & Hat & fit type     \\
        Bracelet & bracelet form & Light Fixture & installation type     \\
        Coat & coat silhouette type & Wall Art & indoor outdoor usage     \\
        \bottomrule
    \end{tabular}
    }
\end{table}

\vspace{-0.8em}
\section{Related Work}
Attribute prediction in e-commerce has evolved from rule-based~\cite{chiticariu2010} and NER-based extraction~\cite{yan2021a} to question-answering reformulations~\cite{yang2022maveqa}. LLMs further expanded the scope by inferring values not explicitly stated in product text~\cite{nikolakopoulos2023sage}, and Zhang et al.~\cite{zhang2025patternrag,zhang2025catalograg,zhang2024titletranslation} improve prediction by leveraging patterns from similar products as contextual guidance. More recently, LLM-based systems have been applied to broader e-commerce tasks: Huang et al.~\cite{huang2025attributeforge} orchestrate specialized LLM agents for automated product schema modeling, Cheng et al.~\cite{cheng2024dualexpert} combine a fine-tuned domain expert with a general LLM for product categorization, Satyadharma et al.~\cite{satyadharma2025autoprompt} develop a training-free LLM cascade for attribute value quality assessment, and Gao et al.~\cite{gao2026cascadeagent} automate prompt specialization for thousands of attribute-specific prompts. Trabelsi et al.~\cite{trabelsi2025vlm} investigate key design considerations for vision-language models applied to product image analysis. These works focus on individual tasks; our work addresses the complementary challenge of continuously monitoring and improving LLM-based attribute prediction through a self-learning manner mediated by Supervisor Agent.

Multi-agent architectures provide structured coordination for complex tasks. Supervisor-tool-calling~\cite{schick2023toolformer,parisi2022talm} and hierarchical variants~\cite{liu2025hmrag} enable agents to leverage external tools and decompose problems. Agent debate and critique have been applied in recommendation~\cite{cai2024flow}, legal~\cite{shengbinyue2025multiagent}, code generation~\cite{hu2025qualityflow}, and general reasoning~\cite{du2023debate} domains. Multi-Expert Prompting~\cite{long2024multiexpert} aggregates role-specific LLM responses to improve factuality. The \emph{LLM-as-a-judge} paradigm~\cite{li2025llmjudge,gu2024llmjudge} offers a cost-effective alternative to human evaluation; in our system, the Evaluator acts as a judge verifying Generator predictions, while the Supervisor serves as a higher-level judge with tool-augmented reasoning.


Recent work explores improving LLMs through feedback and context manipulation rather than weight updates. Self-Refine~\cite{madaan2023selfrefine} uses iterative self-critique, Reflexion~\cite{shinn2023reflexion} maintains verbal reflections in episodic memory, and Malt~\cite{motwani2024malt} improves reasoning through multi-agent training. However, Huang et al.~\cite{huang2023selfcorrect} show that self-correction without external feedback is unreliable — motivating our use of a tool-augmented Supervisor. In automatic prompt optimization, CriSPO~\cite{he2025crispo} applies critique-suggestion modules, DSPy~\cite{khattab2023dspy} compiles declarative modules into optimized pipelines, and ACE~\cite{zhang2025ace} evolves LLM contexts as self-improving playbooks. Our approach differs by mining \emph{disagreement cases} between Generator, Evaluator, and Sellers as targeted improvement signals, aggregating case-level insights into generalizable learnings through the Memory Base and Memory Summarizer, and running the self-learning loop continuously with regression constraints to prevent quality degradation.

\vspace{-0.8em}
\section{Conclusion and Discussion}
\vspace{-0.8em}
We introduce a novel Supervisor Agent architecture within e-commerce catalog enrichment system that intervenes when two types of difficult disagreement cases occurred — Generator-Evaluator LLMs' conflicts, Seller corrections on LLM outputs. The system features a Memory Base and a Memory Summarizer that stores Supervisor Agent activities from individual case processing, and aggregates patterns into learnings that are fed back into worker Generator and Evaluator LLMs to self-improve without human intervention. Future directions include using high quality labels generated from Supervisor Agent to finetune worker Generator and Evaluator LLMs.

\newpage
\appendix
\section{Appendix: Analysis of the Performance of Generator–Evaluator-Supervisor framework}
\label{sec:appendix}

For a given \emph{Product--Attribute} pair $(\texttt{p},\texttt{a})$, let $\mathcal X_G:=(T,I,M^{G}_{\texttt{p},\texttt{a}})$ and $\mathcal X_E:=(T,I,M^{E}_{\texttt{p},\texttt{a}})$ denote the multimodal inputs to the Generator and Evaluator respectively, where $T$ is textual information, $I$ is image information, and $M^\bullet$ is the structured metadata visible to each role. The three pipeline components are:
\[
G_\theta : \mathcal X_G \mapsto (V,R_G),\qquad
E_\phi : \mathcal X_E\times V \mapsto (D,R_E),\qquad
S_\psi : \mathcal X_G\times V\times R_G\times D\times R_E
\mapsto (\widehat V,R_S),
\]
with $D\in\{0,1\}$ (\textsc{Reject}/\textsc{Accept}),
free-text rationales $R_G,R_E,R_S$, and $V\subseteq S$ the attribute's value domain. The catalog publishes $\widehat V$ iff the last decision is \textsc{Accept}. The Supervisor is invoked on slices where the G--E pipeline identified an error. Since the Evaluator never sees $R_G$, we have $(D,R_E)\perp\!\!\!\perp R_G\mid(\mathcal X_E,V)$. This fact keeps the three roles statistically disjoint and will allow us to characterise the Evaluator with exactly two scalars, its true-positive and false-positive rates. Let $Y$ be the (unknown) ground-truth value, we define $G_c:=[V=Y]$, $p_g:=\Pr(G_c=1)$, $a:=\Pr(D=1\mid G_c=1)$ (Evaluator TPR), and $b:=\Pr(D=1\mid G_c=0)$ (Evaluator FPR). The Supervisor intervenes on three \emph{residual slices}:
$\mathcal R_1:(G_c=1,D=0)$ \textbf{recovery}, with traffic mass $M_1:=p_g(1-a)$;\\
$\mathcal R_2:(G_c=0,D=1)$ \textbf{correction}, with traffic mass $M_2:=(1-p_g)b$;\\
$\mathcal R_3:(G_c=0,D=0)$ \textbf{addition}, with traffic mass $M_3:=(1-p_g)(1-b)$.\\
For each slice, the Supervisor produces one of three outcomes: \texttt{ok} (correct value published), \texttt{bad} (incorrect value published), or \texttt{rej} (no value published), with slice-level rates $(p_{\mathrm{rec}}^{\texttt{ok}},p_{\mathrm{rec}}^{\texttt{bad}},p_{\mathrm{rec}}^{\texttt{rej}})$, $(p_{\mathrm{cor}}^{\texttt{ok}},p_{\mathrm{cor}}^{\texttt{bad}},p_{\mathrm{cor}}^{\texttt{rej}})$, $(p_{\mathrm{add}}^{\texttt{ok}},p_{\mathrm{add}}^{\texttt{bad}},p_{\mathrm{add}}^{\texttt{rej}})$, with each triplet summing to $1$. The baseline (time--\textit{t}) precision is
$\operatorname{Prec}_t = {p_g a}/({p_g a + M_2})$,
and the post-Supervisor (time--\textit{t+1}) precision is:                                                                                        
  \begin{equation*}                                                                                                             
    \operatorname{Prec}_{t+1}                                                                                                  
      = \frac{                                                                                                                 
          p_g a                                                                                                                
      + M_1 p_{\mathrm{rec}}^{\texttt{ok}}                                                                                     
      + M_2 p_{\mathrm{cor}}^{\texttt{ok}}                                                                                     
      + M_3 p_{\mathrm{add}}^{\texttt{ok}}}
        {
          p_g a
      + M_1(p_{\mathrm{rec}}^{\texttt{ok}}+p_{\mathrm{rec}}^{\texttt{bad}})
      + M_2(1-p_{\mathrm{cor}}^{\texttt{rej}})
      + M_3(p_{\mathrm{add}}^{\texttt{ok}}+p_{\mathrm{add}}^{\texttt{bad}})}
  \end{equation*}
We are now ready to state the following: inserting a Supervisor after the G--E pair can only improve catalog quality. Every residual slice lifts global precision in proportion to its traffic mass and the Supervisor's \texttt{ok} rate (Theorem~\ref{thm:precision}), while each correct publication from any slice nudges overall accuracy upward and can never decrease it (Theorem~\ref{thm:accuracy}).

\begin{theorem}[Supervisor impact on Precision]
\label{thm:precision}
Define slice-specific precision advantages
$\ell_i := p_{i}^{\texttt{ok}} - \operatorname{Prec}_t$ for $i\in\{\mathrm{rec},\mathrm{cor},\mathrm{add}\}$.
Then
$\Delta^{\mathrm{prec}} := \operatorname{Prec}_{t+1} - \operatorname{Prec}_t > 0
\;\Longleftrightarrow\;
M_1\ell_1 + M_2\ell_2 + M_3\ell_3 > 0$.
\end{theorem}

\begin{corollary}[Sub--$p_g$ Supervisor cannot lift precision]
\label{cor:precision_decay_3slice}
If the Evaluator is non-trivial ($a\ge b$) and $p_{\mathrm{rec}}^{\texttt{ok}},\,p_{\mathrm{cor}}^{\texttt{ok}},\,p_{\mathrm{add}}^{\texttt{ok}} \le p_g$, then $\operatorname{Prec}_{t+1}\le\operatorname{Prec}_t$, with strict inequality when any $M_i>0$ and the corresponding slice precision is $<p_g$.
\end{corollary}

\begin{theorem}[Supervisor impact on accuracy]
\label{thm:accuracy}
Let $A_t:=\Pr[\widehat V=Y]$ be catalog-wide accuracy and $\Delta_t:=A_{t+1}-A_t = M_1 p_{\mathrm{rec}}^{\texttt{ok}} + M_2 p_{\mathrm{cor}}^{\texttt{ok}} + M_3 p_{\mathrm{add}}^{\texttt{ok}}$.
\textbf{(i) Non-regression:} $\Delta_t\ge 0$ for all admissible rate choices.
\textbf{(ii) Strict improvement:} $\Delta_t>0$ iff at least one slice has $M_i>0 \wedge p_i^{\texttt{ok}}>0$.
\end{theorem}


\vspace{-1em}
\begin{thebibliography}{99}

\bibitem{nikolakopoulos2023sage}
Nikolakopoulos, A.N., Kaul, S., Gade, S.K., Dubrov, B., Batur, U., Khan, S.A.: Sage: Structured attribute value generation for billion-scale product catalogs. arXiv preprint arXiv:2309.05920 (2023)

\bibitem{huang2025attributeforge}
Huang, Y., Ramo, K., Iovine, A., Monteiro, M., Gokalp, S., Bakshi, A., Turalic, H., Kumar, A., Neumeier, J., Yates, R., Monir, R., Hartmann, S., Manglik, T., Yakout, M.: AttributeForge: An agentic LLM framework for automated product schema modeling. In: Proceedings of EMNLP 2025, Industry Track (2025)

\bibitem{satyadharma2025autoprompt}
Satyadharma, S., Sheikholeslami, F., Kaul, S., Batur, A.U., Khan, S.A.: Auto prompting without training labels: An LLM cascade for product quality assessment in e-commerce catalogs. In: Proceedings of EMNLP 2025, Industry Track (2025)

\bibitem{gao2026cascadeagent}
Gao, P., Nikolakopoulos, A., Cheng, Z., Scarinci, A., Batur, A.U., Khan, S.A.: Agentic Prompt Optimization for E-Commerce Catalog Enrichment at Scale. arXiv preprint (2026)

\bibitem{li2025llmjudge}
Li, D., Jiang, B., Huang, L., Beigi, A., Zhao, C., Tan, Z., Bhattacharjee, A., Jiang, Y., Chen, C., Wu, T., et al.: From generation to judgment: Opportunities and challenges of llm-as-a-judge. arXiv preprint arXiv:2411.16594 (2025)

\bibitem{gu2024llmjudge}
Gu, J., Jiang, X., Shi, Z., Tan, H., Zhai, X., Xu, C., Li, W., Shen, Y., Ma, S., Liu, H., et al.: A survey on llm-as-a-judge. arXiv preprint arXiv:2411.15594 (2024)

\bibitem{yao2022react}
Yao, S., Zhao, J., Yu, D., Du, N., Shafran, I., Narasimhan, K.R., Cao, Y.: React: Synergizing reasoning and acting in language models. In: The Eleventh International Conference on Learning Representations (2022)

\bibitem{chiticariu2010}
Chiticariu, L., Krishnamurthy, R., Li, Y., Reiss, F., Vaithyanathan, S.: Domain adaptation of rule-based annotators for named-entity recognition tasks. In: Proceedings of EMNLP, pp. 1002--1012 (2010)

\bibitem{yan2021a}
Yan, H., Gui, T., Dai, J., Guo, Q., Zhang, Z., Qiu, X.: A unified generative framework for various NER subtasks. arXiv preprint (2021)


\bibitem{yang2022maveqa}
Yang, L., Wang, Q., Yu, Z., Kulkarni, A., Sanghai, S., Shu, B., Elsas, J., Kanagal, B.: Mave: A product dataset for multi-source attribute value extraction. In: Proceedings of WSDM '22, pp. 1256--1265 (2022)

\bibitem{zhang2025patternrag}
Zhang, B., Khan, S.A., Walter, S.: Leveraging product catalog patterns for multilingual e-commerce product attribute prediction. In: Proceedings of EMNLP 2025, Industry Track (2025)

\bibitem{zhang2025catalograg}
Zhang, B., Khan, S.A., Walter, S.: CatalogRAG: Retrieval-guided LLM prediction for multilingual e-commerce product attributes. In: KDD 2025 Workshop on LLM4ECommerce (2025)

\bibitem{zhang2024titletranslation}
Zhang, B., Walter, S.: Don't just translate, summarize too: Cross-lingual product title generation in e-commerce. In: LREC-COLING 2024 Workshop on e-Commerce and NLP (2024)

\bibitem{cheng2024dualexpert}
Cheng, Z., Zhang, W., Chou, C.-C., Jau, Y.-Y., Pathak, A., Gao, P., Batur, U.: E-commerce product categorization with LLM-based dual-expert classification paradigm. In: Proceedings of the 1st Workshop on Customizable NLP (CustomNLP4U), pp. 294--304 (2024)

\bibitem{trabelsi2025vlm}
Trabelsi, A., Zontak, M., Qian, Y., Jackson, B., Khan, S., Batur, U.: What matters when building vision language models for product image analysis? In: IEEE Winter Conference on Applications of Computer Vision (WACV) Workshops (2025)

\bibitem{schick2023toolformer}
Schick, T., Dwivedi-Yu, J., Dess\`{i}, R., Raileanu, R., Lomeli, M., Hambro, E., Zettlemoyer, L., Cancedda, N., Scialom, T.: Toolformer: Language models can teach themselves to use tools. Advances in Neural Information Processing Systems \textbf{36}, 68539--68551 (2023)

\bibitem{parisi2022talm}
Parisi, A., Zhao, Y., Fiedel, N.: Talm: Tool augmented language models. arXiv preprint arXiv:2205.12255 (2022)

\bibitem{liu2025hmrag}
Liu, P., Liu, X., Yao, R., Liu, J., Meng, S., Wang, D., Ma, J.: Hm-rag: Hierarchical multi-agent multimodal retrieval augmented generation. arXiv preprint arXiv:2504.12330 (2025)

\bibitem{cai2024flow}
Cai, S., Zhang, J., Bao, K., Gao, C., Feng, F.: Flow: A feedback loop framework for simultaneously enhancing recommendation and user agents. arXiv preprint arXiv:2410 (2024)

\bibitem{shengbinyue2025multiagent}
ShengbinYue, S., Huang, T., Jia, Z., Wang, S., Liu, S., Song, Y., Huang, X.-J., Wei, Z.: Multi-agent simulator drives language models for legal intensive interaction. In: Findings of the Association for Computational Linguistics: NAACL 2025, pp. 6537--6570 (2025)

\bibitem{hu2025qualityflow}
Hu, Y., Zhou, Q., Chen, Q., Li, X., Liu, L., Zhang, D., Kachroo, A., Oz, T., Tripp, O.: Qualityflow: An agentic workflow for program synthesis controlled by llm quality checks. arXiv preprint arXiv:2501.17167 (2025)

\bibitem{du2023debate}
Du, Y., Li, S., Torralba, A., Tenenbaum, J.B., Mordatch, I.: Improving factuality and reasoning in language models through multiagent debate. In: Forty-first International Conference on Machine Learning (2023)

\bibitem{long2024multiexpert}
Long, D.X., Yen, D.N., Luu, A.T., Kawaguchi, K., Kan, M.-Y., Chen, N.F.: Multi-expert prompting improves reliability, safety, and usefulness of large language models. arXiv preprint arXiv:2411.00492 (2024)

\bibitem{madaan2023selfrefine}
Madaan, A., Tandon, N., Gupta, P., Hallinan, S., Gao, L., Wiegreffe, S., Alon, U., Dziri, N., Prabhumoye, S., Yang, Y., et al.: Self-Refine: Iterative refinement with self-feedback. In: NeurIPS (2023)

\bibitem{shinn2023reflexion}
Shinn, N., Cassano, F., Gopinath, A., Narasimhan, K., Yao, S.: Reflexion: Language agents with verbal reinforcement learning. In: NeurIPS (2023)

\bibitem{motwani2024malt}
Motwani, S.R., Smith, C., Das, R.J., Rafailov, R., Laptev, I., Torr, P.H.S., Pizzati, F., Clark, R., de Witt, C.S.: Malt: Improving reasoning with multi-agent llm training. arXiv preprint arXiv:2412.01928 (2024)

\bibitem{huang2023selfcorrect}
Huang, J., Chen, X., Mishra, S., Zheng, H.S., Yu, A.W., Song, X., Zhou, D.: Large language models cannot self-correct reasoning yet. In: ICLR (2024)

\bibitem{he2025crispo}
He, H., Liu, Q., Xu, L., Shivade, C., Zhang, Y., Srinivasan, S., Kirchhoff, K.: CriSPO: Multi-aspect critique-suggestion-guided automatic prompt optimization for text generation. In: Proceedings of the AAAI Conference on Artificial Intelligence, vol. 39, no. 22, pp. 24014--24022 (2025)

\bibitem{khattab2023dspy}
Khattab, O., Singhvi, A., Maheshwari, P., Zhang, Z., Santhanam, K., et al.: DSPy: Compiling declarative language model calls into self-improving pipelines. In: ICLR (2024)

\bibitem{zhang2025ace}
Zhang, Q., Hu, C., Upasani, S., Ma, B., Hong, F., Kamanuru, V., Rainton, J., Wu, C., Ji, M., Li, H., Thakker, U., Zou, J., Olukotun, K.: Agentic context engineering: Evolving contexts for self-improving language models. arXiv preprint arXiv:2510.04618 (2025)









\end{thebibliography}
\end{document}